\documentclass[runningheads]{llncs}
\usepackage[T1]{fontenc}
\usepackage{graphicx}
\usepackage{url}
\usepackage{booktabs}  
\usepackage{amsfonts}  
\usepackage{nicefrac} 
\usepackage{microtype}
\usepackage{lipsum}
\usepackage{amsmath}
\usepackage{graphicx}
\usepackage{xcolor}
\usepackage{multirow}
\usepackage{enumitem}
\usepackage{tabularx}
\usepackage{subcaption}
\usepackage{floatrow}
\usepackage{gensymb}
\usepackage{tabularray}
\usepackage{float}
\usepackage{floatrow}
\usepackage{soul}

\begin{document}
\title{A Novel Approach for Pill-Prescription Matching with GNN Assistance and Contrastive Learning}
\titlerunning{Pill-Prescription Matching with GNN Assistance and Contrastive Learning}
\author{Trung Thanh Nguyen\inst{1}, Hoang Dang Nguyen\inst{1}, Thanh Hung Nguyen\inst{1},\\ Huy Hieu Pham\inst{2,3}, Ichiro Ide\inst{4}, Phi Le Nguyen\inst{1}\inst{*}}

\authorrunning{Trung Thanh Nguyen et al.}

\institute{\inst{1}School of Information and Communication Technology, Hanoi University of Science and Technology, Vietnam; \\\{thanh.nt176874@sis, dang.nh194423@sis, hungnt@soict, lenp@soict\}.hust.edu.vn \\
\inst{2}College of Engineering \& Computer Science, VinUniversity, Hanoi, Vietnam; hieu.ph@vinuni.edu.vn\\
\inst{3}VinUni-Illinois Smart Health Center, VinUniversity, Hanoi, Vietnam\\
\inst{4}Graduate School of Informatics, Nagoya University, Japan; ide@i.nagoya-u.ac.jp\\
\inst{*}Corresponding author
}
\maketitle 
\vspace{-10pt}
\begin{abstract}
Medication mistaking is one of the risks that can result in unpredictable consequences for patients. To mitigate this risk, we develop an automatic system that correctly identifies pill-prescription from mobile images. Specifically, we define a so-called pill-prescription matching task, which attempts to match the images of the pills taken with the pills' names in the prescription. We then propose PIMA, a novel approach using Graph Neural Network (GNN) and contrastive learning to address the targeted problem. In particular, GNN is used to learn the spatial correlation between the text boxes in the prescription and thereby highlight the text boxes carrying the pill names. In addition, contrastive learning is employed to facilitate the modeling of cross-modal similarity between textual representations of pill names and visual representations of pill images. We conducted extensive experiments and demonstrated that PIMA outperforms baseline models on a real-world dataset of pill and prescription images that we constructed. Specifically, PIMA improves the accuracy from $19.09\%$ to $46.95\%$ compared to other baselines. We believe our work can open up new opportunities to build new clinical applications and improve medication safety and patient care. 

\end{abstract}
\section{Introduction}

A WHO report states that drug abuse, rather than illness, accounts for one-third of all deaths \cite{8679954}.
Additionally, roughly $6,000$ -- $8,000$ persons per year pass away due to drug errors, according to Yaniv et al. \cite{8010584}.
Medical errors could seriously damage the treatment effectiveness, cause unfavorable side effects, or even lead to death.
WHO has picked the theme Medication Without Harm for World Patient Safety Day 2022 to highlight the significance of taking medication properly.
Drug abuse can be brought on by various factors, including using pills other than those prescribed. To this end, this study concentrates on the issue of matching the pill names in a prescription to the corresponding pills in an image, thereby detecting missing or mistaken pills. We call this the \emph{pill-prescription matching} problem.
The problem's context can be described as follows.
The user has a prescription image and an image capturing pills that will be taken. We want to match each pill in the pill image with the corresponding name in the prescription. 

The pill-prescription matching task is analogous to the well-known text-image matching task. In the text-image matching task, an input consists of an image containing numerous objects and a short paragraph of sentences. The text-image matching task aims to identify the keywords in the sentences and match them with the related objects in the image. The key issue in text-image matching is measuring the visual-semantic similarity between a text and an image. Frome et al. \cite{fromedeep} proposed a feature embedding framework that uses Skip-Gram and Convolutional Neural Network (CNN) to extract cross-modal feature representations. Then, the ranking loss is applied so that the distance between mismatched text-image pairs is greater than that between matched pairs. Kiros et al. \cite{kiros2014unifying} utilized a similar approach that leverages the Long Short-Term Memory (LSTM) \cite{hochreiter1997long} to generate text representations. With the recent success of pre-training and self-supervised learning, text-image matching has profited from the rich visual and linguistic representation of pre-trained models on large-scale datasets for downstream tasks. Radford et al. \cite{radford2021learning} proposed a Contrastive Language-Image Pre-Training (CLIP) model to learn visual concepts under language supervision. It is trained using 400 million (text, image) pairs collected from the Web. Gao et al. \cite{gao2020fashionbert} examined the text-image matching in cross-modal retrieval of the fashion industry. They used the pre-trained Bidirectional Encoder Representations from Transformers (BERT) \cite{devlin2018bert} as the backbone network to learn high-level representations of texts and images.

However, the pill-prescription matching task differs from the general text-image matching task in the following aspects.
First, unlike the common text-image matching task, pill names are typically lengthy phrases (instead of words like in the general text-image matching problem). Notably, the pill's name has almost no semantic meaning.
Additionally, the same pill name might be expressed in a variety of ways (depending on the doctors).
Moreover, many text boxes in the prescription (e.g., quantities, diagnostic, etc.) do not relate to the pill name. 
Therefore, the conventional text-image matching approaches are inadequate for the pill-prescription matching issue. To this end, we propose a novel approach for dealing with the pill-prescription matching problem. Our main idea is to leverage a Graph Neural Network (GNN) for capturing the spatial relationship of text boxes in the prescription, thereby highlighting text boxes that contain pill names. 
Moreover, we propose a cross-modal matching mechanism that employs a contrastive loss to encourage the distance between the mismatched pill image and pill name pairs while minimizing that of the matched pairs.

In summary, this work makes the following contributions:
\vspace{-3pt}
\begin{enumerate}
\item We propose PIMA, a novel deep learning framework based on GNN and contrastive learning for the pill-prescription matching problem. To the best of our knowledge, we are the first to define and address this challenging task on a real-world dataset. The method is applicable for real-world scenarios in clinical practice to improve medication safety.
\item We conduct comprehensive experiments to demonstrate the effectiveness of the proposed approach on a real-world pill-prescription dataset. The proposed PIMA outperforms baseline methods with significant improvements in performance.
\end{enumerate}

The remainder of the paper is organized as follows. 
We briefly introduce GraphSAGE, a GNN used in our proposed model in Section \ref{sec:preliminaries}. 
The details of our proposed method are described in Section \ref{sec:proposal}.
We perform experiment to evaluate the proposal in Section \ref{sec:evaluation} and conclude the paper in Section \ref{sec:conclusion}.


\vspace{-5pt}
\section{Preliminaries}
\label{sec:preliminaries}
\begin{figure}[tb]
    \centering
    \includegraphics[width=0.6\textwidth]{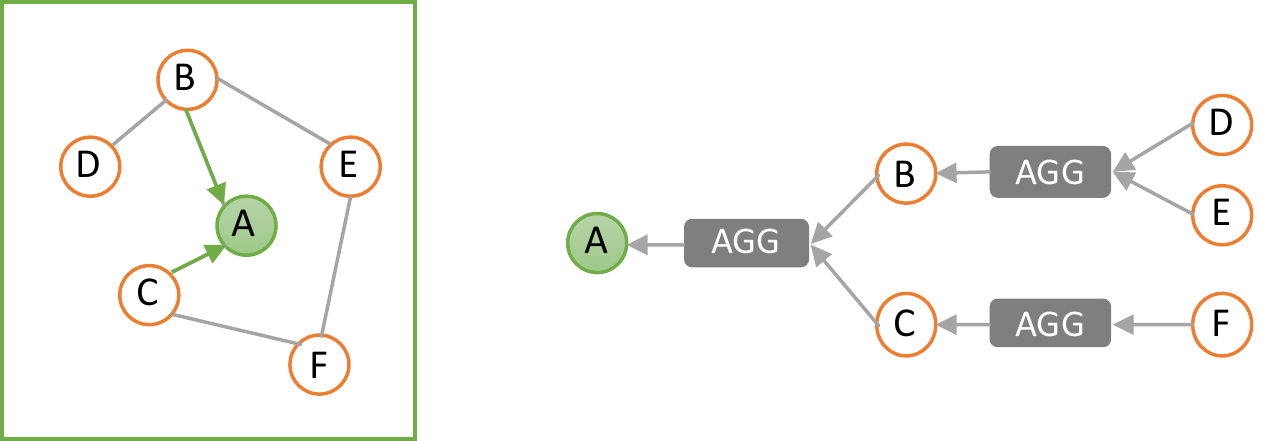}
    \caption{\small{The left side represents a graph, whereas the right side depicts the aggregation process of GraphSAGE (with two convolution layers) to generate the embedding vector for vertex $A$.}\vspace{-10pt}}
    \label{fig:agg_graph}
\end{figure}
This section introduces \textit{Graph SAmple and aggreGatE} (GraphSAGE), which is one of the most well-known Graph Neural Network (GNN) developed by Hamilton et al. \cite{hamilton2017inductive}.
Similar to the convolution operation in Convolutional Neural Network (CNN), in the GraphSAGE, information relating to the local neighborhood of a node is collected and used to compute the node embeddings. For each node, the algorithm iteratively aggregates information from its neighbors.
At each iteration, the neighborhood of the node is initially sampled, and the information from the sampled nodes is aggregated into a single vector.
Specifically, at the $k$-th layer, the aggregated information $h^k_{N(v)}$ at a node $v$, based on the sampled neighborhood $N(v)$, can be expressed as:
\begin{equation}
    h^k_{N(v)} = \text{AGG}_k\left( \left\{ h_u^{k-1}, \forall{u} \in N(v) \right\} \right), \nonumber
\end{equation}
where $h_u^{k-1}$ is the embedding of node $u$ in the previous layer. 
The aggregated embeddings of the sampled neighborhood $h^k_{N(v)}$ then is concatenated with the node's embedding at the previous layer $h_v^{k-1}$ to form its embedding at the current layer as follows.
\begin{equation}
    h_v^k = \sigma \left( W^k \cdot \text{CONCAT} \left( h_v^{k-1}, h_{N(v)}^k \right) \right), \nonumber
    \label{eq:node_concat}
\end{equation}
where $W^k$ is a trainable weight matrix and $\sigma$ is a non-linear activation function.
Figure~\ref{fig:agg_graph} illustrates the aggregation process of GraphSAGE.
GraphSAGE offers several aggregation methods, including mean, pooling, or neural networks (e.g. Long Short-Term Memory (LSTM) \cite{hochreiter1997long}).


\vspace{-5pt}
\section{Proposed Method}

\label{sec:proposal}
This section describes our proposed method named PIMA for the \textbf{PI}ll-prescription \textbf{MA}tching problem. We start by providing an overview of our solution in Section~\ref{subsec:overview}. We then describe the Pill Detector and Prescription Recognizer modules in Sections \ref{subsec:pill_image} and \ref{subsec:prescription}, respectively. 
Finally, we explain the proposed loss function in Section \ref{subsec:loss}.

\vspace{-5pt}
\subsection{Overview}
\label{subsec:overview}
\begin{figure}[tb]
    \centering
    \includegraphics[width=1.0\textwidth]{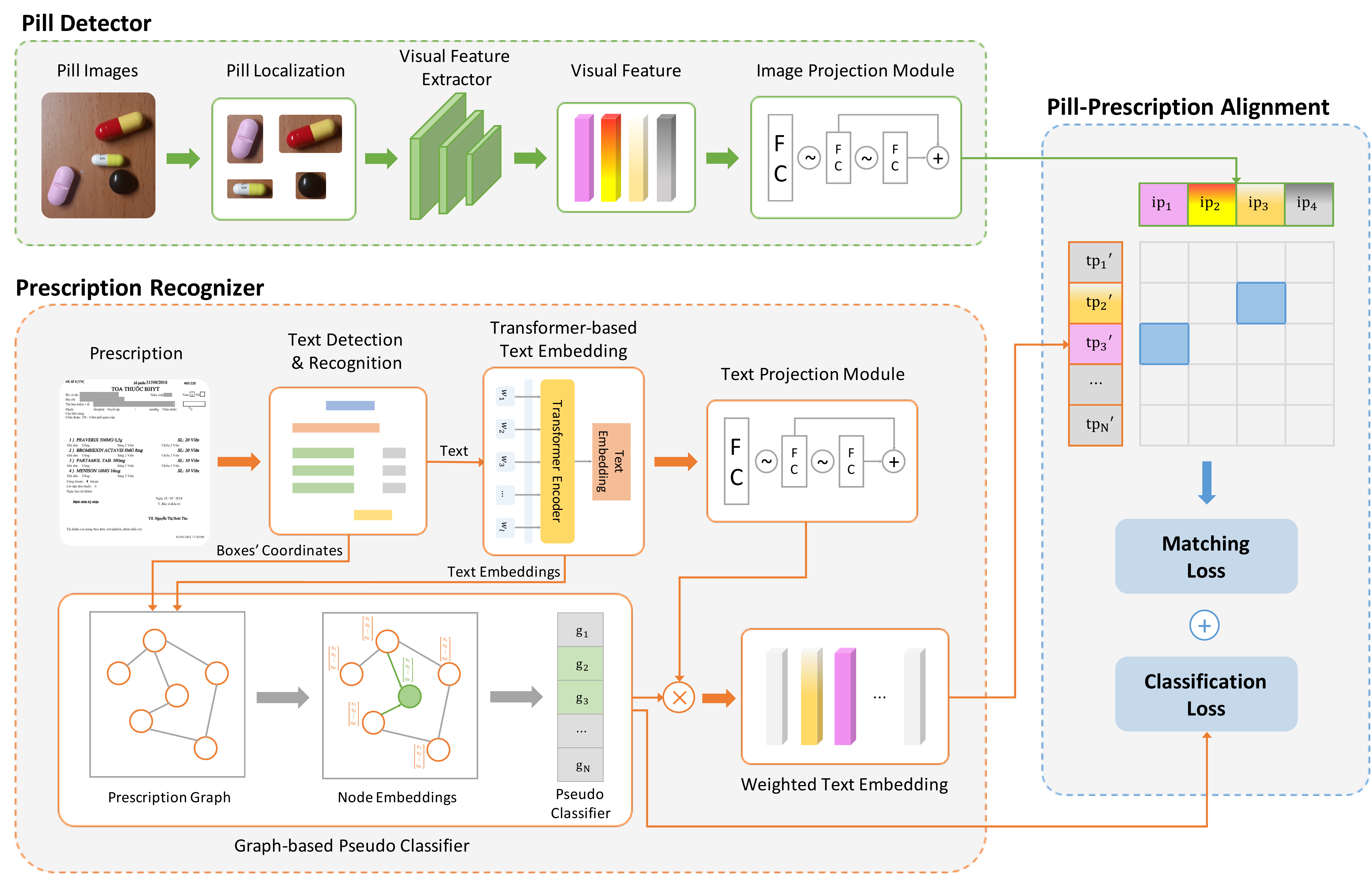}
    \caption{\small{Overview of PIMA, which consists of three modules: Pill Detector, Prescription Recognizer, and Pill-Prescription Alignment. The Pill Detector is responsible for extracting visual features from pill images. The Prescription Recognizer is responsible for embedding each text box on the prescription and highlighting those with pill names. Finally, the textual and visual data are fed into Pill-Prescription Alignment to produce a matching result.\vspace{-10pt}
    }}
    \label{fig:proposed_framework}
\end{figure}
Figure \ref{fig:proposed_framework} illustrates the overview of our proposed model, which consists of three modules. The first module, named \textbf{Pill Detector}, uses Convolutional Neural Network (CNN) to create representations of pills. The second module, named  \textbf{Prescription Recognizer}, is responsible for extracting the textual information. Specifically, this module utilizes a transformer model to create embedding of the text boxes. Moreover, we leverage a Graph Neural Network (GNN) to capture spatial information among the text boxes, thereby highlighting boxes representing the pill names. Finally, the textual and visual features are projected into a shared space before being fed into a Pill-Prescription Alignment module. In the \textbf{Pill-Prescription Alignment} module, we employ two loss functions. The first loss, a Binary Cross-entropy loss, is responsible for classifying the text boxes containing the pill name, while the second loss, a Contrastive loss, is responsible for matching the pill names with the associated pill images.

\vspace{-5pt}
\subsection{Pill Detector}
\label{subsec:pill_image}
The Pill Detector consists of two main components: Pill Localization and Feature Extraction. 
Firstly, an object detection model is applied to determine the location of every pill. 
Assuming that there are $M$ pills cropped from an image, let's denote them as $\left \{ p_1, ..., p_M \right \}$. 
We then leverage a CNN to extract visual features of the pills and obtain $M$ feature vectors $\left \{ \textbf{ie}_1, ..., \textbf{ie}_M \right \}$. 
These feature vectors are then projected onto the same hyper-plane with their counterpart in the prescription via an Image projection module. Consequently, we come up with the final representation of the pills as $\textbf{IP} = \left [ \textbf{ip}_1, ..., \textbf{ip}_M\right ]$.

\vspace{-5pt}
\subsection{Prescription Recognizer}
\label{subsec:prescription}
The Prescription Recognizer comprises three sub-modules: Text Recognition, Text Embedding, and Pseudo-Classification.  
Initially, the text recognition model is utilized to identify and localize each text box in the prescription.
Suppose the prescription contains $N$ text boxes, denoted as $\{s_1, ..., s_N\}$,
then they are put through a Transformer-based text embedding module~\cite{vaswani2017attention} to produce embedding vectors of the text boxes. 
On the one hand, these embedding vectors, along with the coordinates of the text boxes, are utilized to construct a graph representing the spatial relationship between the text boxes. This graph is used as the pseudo-classifier's input to highlight the boxes containing the pill names. On the other hand, these embedding vectors capture contextual information of the text boxes, which is used for matching with pill images.


\textbf{\\Transformer-based Text Embedding.}
Given a text box $s_i = [w_1^{(i)}, ..., w_{l_i}^{(i)}]$, where $w_t^{(i)}$ ($t=1, ..., l_i$) represents the token embedding of the $t$-th word of $s_i$, the text embedding of $s_i$, denoted by $\textbf{te}_{i}$, is obtained by feeding $[w_1^{(i)}, ..., w_{l_i}^{(i)}]$ into a Transformer encoder.
These text embeddings are then projected to the same hyper-plane as their counterparts in the pill images with the aid of a text projection module. 
Consequently, we get the final representations of the $N$ text boxes as $\textbf{TP} = \left[\textbf{tp}_1, \cdots, \textbf{tp}_N \right]$.


\textbf{\\Graph-based Pseudo Classifier.} We noticed that a prescription contains numerous text boxes without pill names.
Therefore, to improve the pill-prescription matching accuracy, a preprocessing step is required to highlight the boxes that are likely to contain pill names.
For this, we create a binary classifier based on a GNN.
Specifically, we first construct a graph $G$ representing the spatial relationship between the text boxes, i.e., $G=\{V, E\}$, where $V =\{v_1, ..., v_N\}$, with $v_i$ the $i$-th text box. The attribute of $v_i$ is the text embedding $\textbf{te}_i$.
Two vertices, $v_i$ and $v_j$, are connected if either $v_i$ or $v_j$ is the box with the shortest horizontal (or vertical) distance to the other. 
Any network can be used for this purpose, and investigating GNN is out of the scope of this paper. 
In this work, we utilize one of the most prominent GNN, namely GraphSAGE~\cite{hamilton2017inductive}, to convert from graph space to vector space.
Consequently, for every vertex $v_i$, we obtain a graph embedding vector $h_i$, which represents the relationship between $v_i$ and its $K$-hop neighbors.
Finally, these graph embedding vectors will then be input to a sigmoid layer to produce the classification results.
In particular, the output of the pseudo-classifier is a vector $g=\left \langle g_1, ...g_N \right \rangle$, where $g_i$ represents the likelihood that the $i$-th textbox has the pill name.
This pseudo-classifier will be trained via classification loss (see Section \ref{subsec:loss}).

Finally, the pseudo classification result is multiplied with the text embedding to obtain the weighted version, $\textbf{TP}' = \left [g_1 \times \textbf{tp}_1, ..., g_N \times \textbf{tp}_N  \right ]$. It is worth noting that as $g_i$ quantifies the likelihood that the $i$-th text box includes the pill name, the embedding vectors of the boxes having the pill name will be highlighted in the $\textbf{TP}'$, while the remainder will be grayed out. 
\vspace{-5pt}
\subsection{Pill-Prescription Alignment}
\label{subsec:loss}
This module receives visual representations of pill images from the Pill Detector and textual representations of text boxes from the Prescription Recognizer. It matches pill names and pill images to generate the final result.
For this, we design an objective function consisting of two losses: classification loss and matching loss.

\textbf{\\Classification Loss.}
We adopt Binary cross-entropy loss to identify whether or not a text box contains a pill name.
We observe that the number of pill name boxes is significantly smaller than that of boxes without pill names.
For this reason, we employ the following weighted cross-entropy loss to mitigate the bias.

\begin{equation}
    \mathcal{L_\text{Classification}} = -\frac{1}{N} \sum_{i = 1}^{N} w_i\left[ y_i\text{log}(g_i) + (1-y_i) \text{log}(1 - g_i) \right], \nonumber
\end{equation}
where $y_i$ and $g_i$ indicate the ground-truth label and the predicted result concerning a text box $s_i$, respectively; $w_i$ represents the ratio of boxes with the label of $1-y_i$.
To be more specific, let $n_1$ be the number of text boxes containing a pill name, and $N$ be the total number of the text boxes, then we have 

\begin{equation}
   w_i = \left\{\begin{matrix}
 1 - \frac{n_1}{N} & \text{,} & \text{ ~if the text box~} s_i \text{~contains a pill name,} \\
 \frac{n_1}{N} &  \text{,}& \text{~otherwise}.
\end{matrix}\right. \nonumber
\end{equation}
\noindent \textbf{Matching Loss.}
This loss aims to model the cross-modal similarity of pill name boxes and pill images' representations.
The principle is to encourage the distance between the representations of mismatched pill names and pill image pairs while minimizing the gap between that of matched pairs. 
Specifically, let $\textbf{ip}_i$ and $\textbf{tp}_j$ be the representations of a pill image $p_i$ and a text box $s_j$, respectively, then their similarity is defined by the cosine similarity as follows. 

\begin{equation}
S(\textbf{ip}_i, \textbf{tp}_j) = \frac{\textbf{ip}_i \cdot \textbf{tp}_j}{\text{max}(\left\| \textbf{ip}_i \right\|_2 \cdot \left\| \textbf{tp}_j \right\|_2, \epsilon)}, \nonumber
\end{equation}
where $\epsilon$ is a small value which is responsible for avoiding division by zero.
The matching loss is then defined as the sum of dis-similarities over all the matched pairs and similarities over all the mismatched pairs as follows. 

\begin{equation*}
    \mathcal{L}_{\text{Matching}} = \frac{1}{M} \sum_{i = 1}^{M} \left [\frac{1}{2}\sum_{j \notin \mathbb{P}_i} S(\textbf{ip}_i, \textbf{tp}_j)^2 + \frac{1}{2}\sum_{k \in \mathbb{P}_i} \text{max}\left (0,m - S\left (\textbf{ip}_i, \textbf{tp}_k \right) \right )^2\right ],
\end{equation*}
where $\mathbb{P}_i$ is the set of all text boxes containing the pill name corresponding to the pill $p_i$, and $m$ is a positive margin specifying the radius surrounding $S\left (\textbf{ip}_i, \textbf{tp}_k \right)$. In our method, we set $m$ to $1$ as the similarity values range from $-1$ to $1$.

The total loss then is defined by the sum of the classification loss and the matching loss as follows.

\begin{equation}
    \mathcal{L}_{\text{Total}} = \mathcal{L}_{\text{Matching}} + \lambda\mathcal{L}_{\text{Classification}},
    \label{eq:loss_function}
\end{equation}
where $\lambda$ is a hyper parameter that balances these two losses.




\vspace{-5pt}
\section{Experiments}
\label{sec:evaluation}
In this section, we conduct comprehensive experiments to evaluate the proposed approach, PIMA. We carefully compare it to the state-of-the-art (SOTA) text-image matching methods under the same experimental settings. Moreover, we perform extensive ablation studies to provide a deeper understanding of some key properties of PIMA.
\vspace{-5pt}
\subsection{Dataset and Experimental Setup}
\label{subsub:dataset}
\begin{figure}[tb]
  \centering
  \includegraphics[width=1.\linewidth]{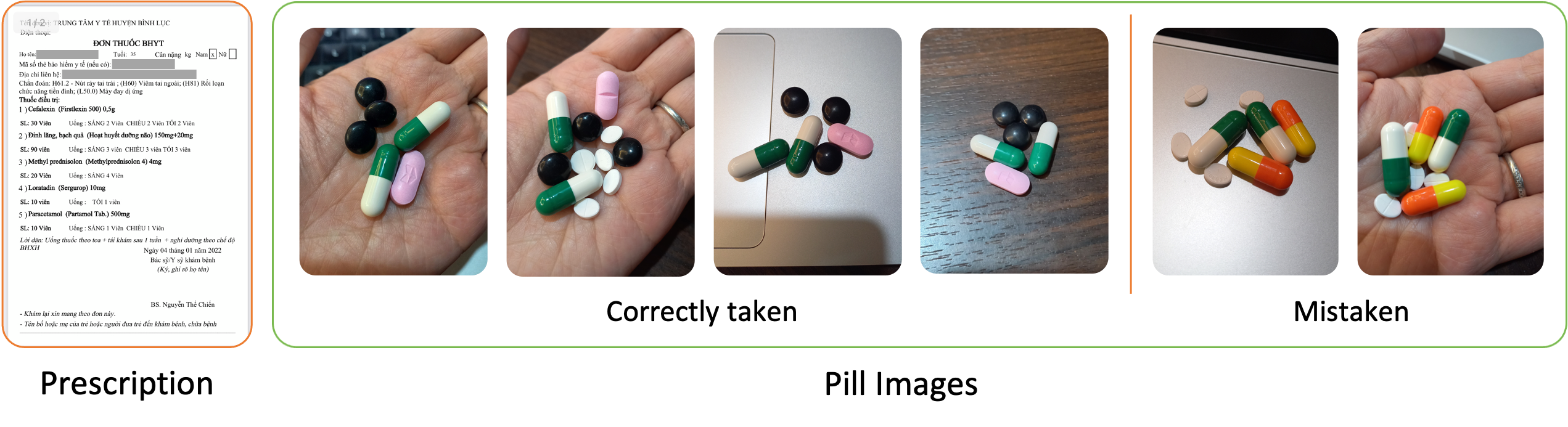}
  \caption{\small{Representative examples from the pill image dataset. It was collected in real-world scenarios, where samples are taken in unconstrained environments.}}
  \label{fig:dataset-2}
\end{figure}

To the best of our knowledge, there is currently no dataset publicly available for the pill-prescription matching task. This motivates us to build an open large-scale dataset containing pill images and the corresponding prescriptions\footnote{The dataset can be downloaded from our project Web-page at \url{https://vaipe.org/#resource}.}. In particular, we collected $1,527$ prescriptions from anonymous patients in major hospitals between 2021 and 2022. After carefully checking data against the privacy concern, we performed the annotation process in which each pill image was assigned and annotated by a human annotator. We then separated the medication intakes for each prescription into morning, noon, and evening parts. For each pill intake, we took about $5$ pictures of the pills. Consequently, we collected $6,366$ pictures of pills, and the unique number of pills was $107$. Figure~\ref{fig:dataset-2} shows several representative examples from our pill dataset. For algorithm development and evaluation, we divided the prescriptions and the corresponding pills into two subsets for training and testing. Details are described in Table~\ref{tab:train_test_rate}.

\textbf{\\Experimental Scenarios.} We evaluate the performance of the proposed PIMA in two distinct circumstances. In the first scenario, we consider settings where medications are taken exactly as prescribed. We assess the precision with which our algorithm could assist users in matching the pills they have taken with their prescription names. The second scenario refers to circumstances in which the pills consumed do not correspond to the prescription. In such a scenario, we examine the accuracy with which our algorithm can identify pills that have been improperly used. Specifically, the percentage of pills incorrectly taken in the second scenario is set to $50\%$. To better investigate the performance of the proposed approach on different data distributions, we divide the two scenarios into sub-scenarios 1-1, 1-2, 1-3, 2-1, and 2-2, which are described in detail in the following.
\begin{table}[tb]
\begin{tblr}{
  colspec = {p{3.3cm}|X[c]|X[c]|X[c]|X[c]},
}
\hline
\SetCell[r=2]{l}\textbf{Experimental Scenario} & \SetCell[c=2]{c}\textbf{Prescription Images} && \SetCell[c=2]{c}\textbf{Pill Images}\\
\cline{2-5} 
 & Train (\%) & Test (\%) & Train (\%) & Test (\%) \\ \hline
Scenario 1-1 & 69.55 & 30.45 & 70.58 & 29.42 \\\hline
Scenario 1-2 & 38.89 & 61.11 & 45.04 & 54.96 \\ \hline
Scenario 1-3 & 3.86 & 96.14 & 6.12 & 93.88 \\ \hline
Scenario 2-1, 2-2 & 69.55 & 30.45 & 72.40 & 27.60 \\ \hline
\end{tblr}
\caption{\small{Details of the data partition.}}
\label{tab:train_test_rate}
\vspace{-10pt}
\end{table}

\begin{figure}[tb]
  \begin{subfigure}[b]{0.45\textwidth}
    \includegraphics[width=\textwidth]{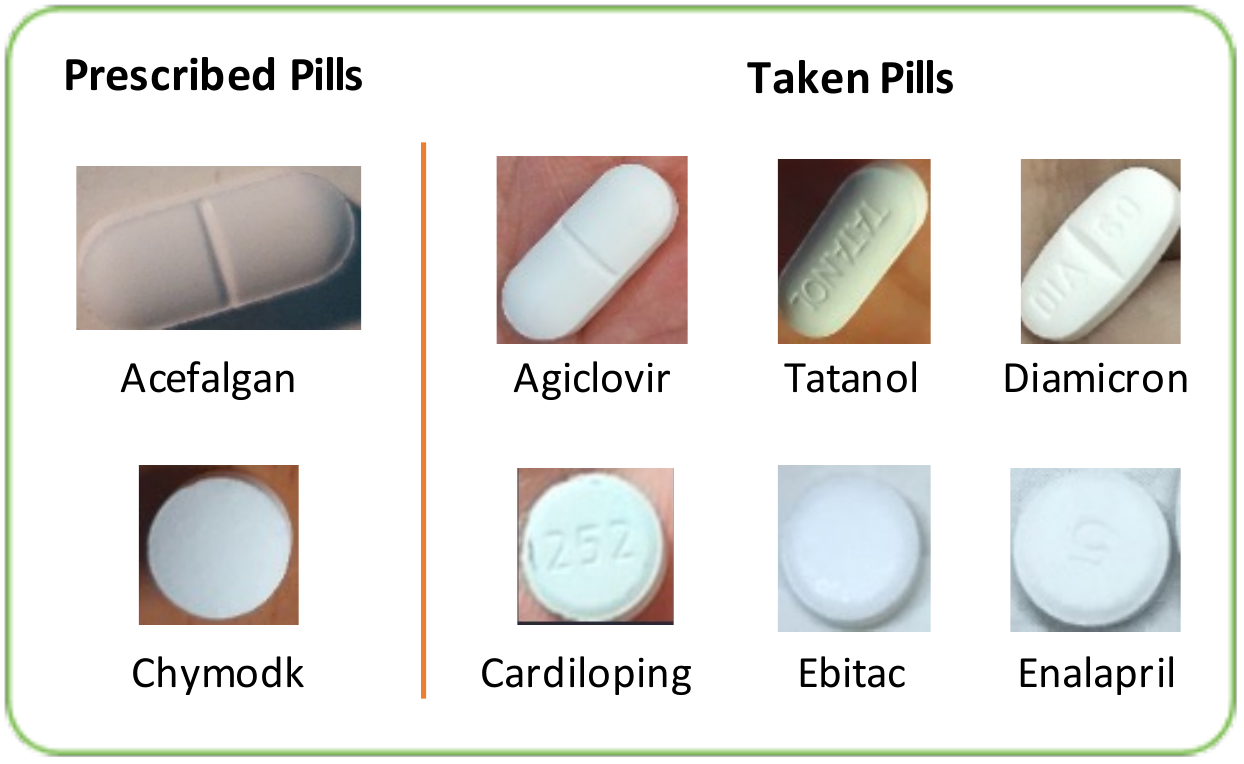}
    \caption{\small{Illustration for Scenario 2-1.}}
    \label{fig:change_exp04}
  \end{subfigure}
  \hfill
  \begin{subfigure}[b]{0.45\textwidth}
    \includegraphics[width=\textwidth]{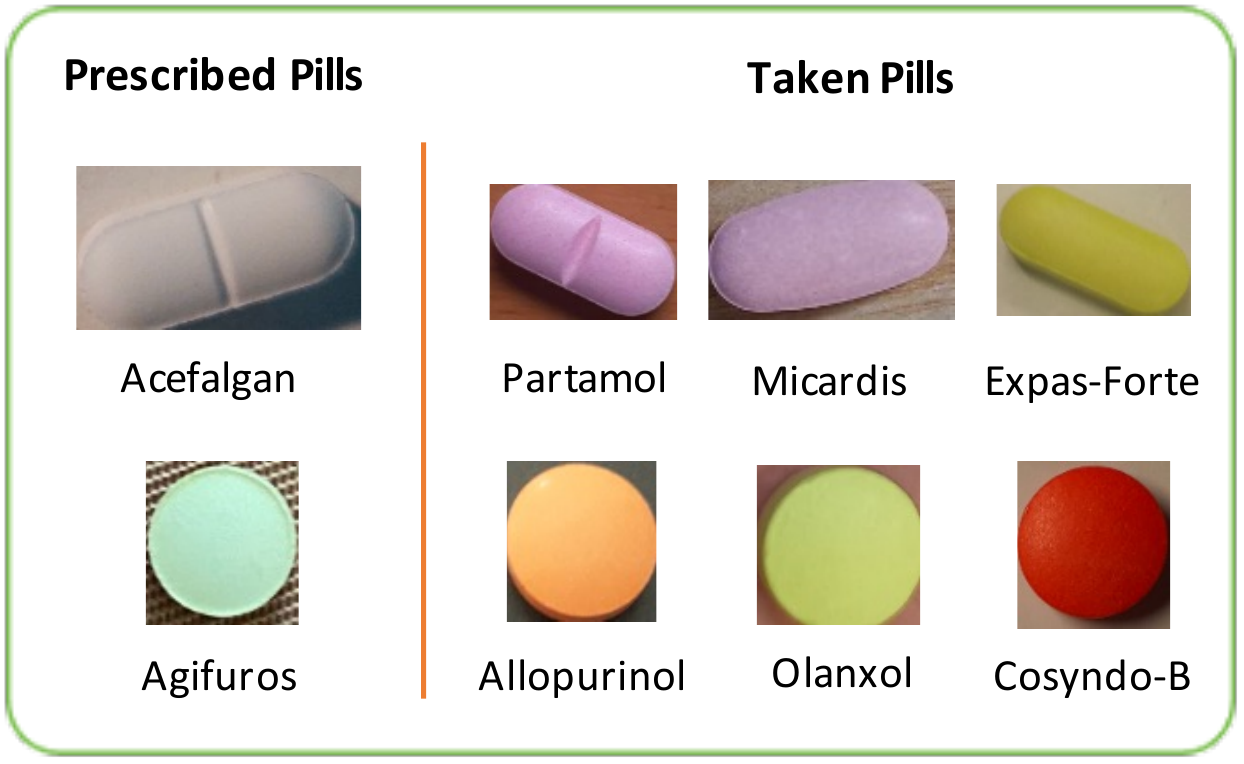}
    \caption{\small{Illustration for Scenario 2-2.}}
    \label{fig:change_exp05}
  \end{subfigure}
  \caption{\small{Visualization of several changes in experimental Scenarios 2-1 and 2-2.}\vspace{-10pt}}
  \label{fig:change_exp_0405}
\end{figure}

\textbf{\\\textit{Scenario 1:}}
\begin{itemize}
    \item[$\bullet$] Scenario 1-1. To split the prescription dataset where each prescription has multiple pill names, we use the stratified sampling method proposed by Sechidis et al. \cite{sechidis2011stratification} into two datasets for training and testing.
    \item[$\bullet$] Scenario 1-2. The prescription data is split so that the training set only has prescriptions that do not overlap, and the test set has all the remaining data. 
    \item[$\bullet$] Scenario 1-3. The prescription data is split so that a pill name only appears once in the training set, and the test set has all the remaining data. 
\end{itemize}
\textbf{\textit{Scenario 2:}} The training dataset is identical to the train set in Scenario 1-1; however, the test set in Scenario 2-1 contains $50\%$ of random pill images that are very similar to the pills in the prescription (in terms of both color and shape); and the test set in Scenario 2-2 contains $50\%$ of random pill images that have the same shape but a different color than the pills in the prescription. Figure~\ref{fig:change_exp_0405} illustrates some examples.

\textbf{\\Evaluation Metrics.} To evaluate the effectiveness of PIMA, we report the final test accuracy using \textit{F1-score} as the main metric, which is widely used in recognition tasks.

\textbf{\\Training Details.} In our implementation, the projection modules consist of two fully connected layers with the Gaussian Error Linear Units (GELU) activation \cite{hendrycks2016gaussian}. The output dimension is set to $256$.
We use AdamW \cite{loshchilov2017decoupled} as the optimizer and set the initial learning rate to $0.001$. The factor $\lambda$ (Eq.~\ref{eq:loss_function}) is set as $1$ for simplicity.  We train the model with the batch size of $4$ and the input image size of $224 \times 224$ pixels. The random rotation of $10^\mathrm{o}$ and horizontal flip are used as data augmentation techniques. All implementations are performed using the PyTorch framework, and the training process is conducted on a machine with an NVIDIA GeForce RTX 3080 GPU.

\vspace{-5pt}
\subsection{Experimental Results}
\label{sub:exp_baseline}
We provide in this section our experimental results and the comparison between our approach and
baseline methods. Because there is no other end-to-end method like ours, the baselines are performed independently in a two-step process. A deep learning-based pill detector is firstly trained to recognize pills from images. The predicted pills are then compared with the correctly identified pill name in the prescription.
\begin{table}[tb]
\begin{tblr}{
  colspec = {p{3.4cm}|p{1.9cm}|X[c]X[c]|X[c]X[c]},
}
\hline
\SetCell[r=2]{l}{\textbf{CNN Model}} & \SetCell[r=2]{c}\SetCell[c=1]{c}{\textbf{Number of Parameters}} & \SetCell[c=2]{c}{\textbf{Pretrained}} && \SetCell[c=2]{c}{\textbf{Non-Pretrained}}\\
\cline{3-6} 
& & Train {\fontsize{8pt}{8pt}\selectfont (\%)} & Test {\fontsize{8pt}{8pt}\selectfont (\%)} & Train {\fontsize{8pt}{8pt}\selectfont (\%)} & Test {\fontsize{8pt}{8pt}\selectfont (\%)} \\ \hline
Resnet-18 \cite{he2016deep} & \SetCell[c=1]{r}11.7M & 99.87 & 86.06 & 67.39 & 67.27 \\  
Resnet-34 \cite{he2016deep} & \SetCell[c=1]{r}21.8M & 96.28 & 82.27 & 68.01 & 66.68 \\  
Resnet-50 \cite{he2016deep} & \SetCell[c=1]{r}25.6M & 86.08 & 77.04 & 68.39 & 67.27 \\  
ViT-Small/16 \cite{dosovitskiy2020image} & \SetCell[c=1]{r}22.1M & 99.44 & 66.47 & 61.07 & 56.59 \\ 
MobileNet-V2 \cite{sandler2018mobilenetv2} & \SetCell[c=1]{r}3.5M & 99.39 & 89.48 & 75.96 & 70.63 \\ 
MobileNet-V3-small \cite{DBLP:journals/corr/abs-1905-02244} & \SetCell[c=1]{r}2.5M & 99.98 & \textbf{90.18} & 74.78 & \textbf{71.65} \\ 
MobileNet-V3-large \cite{DBLP:journals/corr/abs-1905-02244} & \SetCell[c=1]{r}5.5M & 99.93 & 90.01 & 70.31 & 70.30 \\ \hline
\end{tblr}
\caption{\small{Experimental results on the test set of our pill image dataset with different CNN models. The best results are highlighted in \textbf{bold}.\label{tab:experiment_cnn}}}
\end{table}
\vspace{-5pt}
\noindent \textbf{\\ \\Baseline Performance.} In this experiment, we train a set of SOTA Convolutional Neural Network (CNN) architectures (see Table~\ref{tab:experiment_cnn}) on our pill image dataset, in which the information from the prescription is not taken into account. Our experiments follow the Scenario 1-1 setting. During the training process, each learning model is initialized with pre-trained weights on the ImageNet dataset~\cite{deng2009imagenet}. We also investigate the learning performance of CNN models on the pill image dataset when training from scratch. We obtain the highest performance with \texttt{Mobilenet-V3-small} model. It reports an F1-score of $90.18\%$ when using ImageNet-trained deep features and an F1-score of $71.65\%$ when trained from scratch, respectively.
\begin{table}[tb]
\begin{tblr}{
  colspec = {p{5.2cm}|X[c]|X[c]X[c]},
}
\hline
\SetCell[r=2]{l}{\textbf{BERT Model}} & \SetCell[r=2]{c}{\textbf{Task}} &  \SetCell[c=2]{c}{\textbf{CNN Non-Pretrained}}\\
\cline{3-6} 
& & Train {\fontsize{8pt}{8pt}\selectfont (\%)} & Test {\fontsize{8pt}{8pt}\selectfont (\%)} \\ \hline
BERT-base-uncased \cite{devlin2018bert} &  \SetCell[r=2]{c}{Fill-Mask}  & 97.57 & 76.16  \\ 
BERT-base-uncased-multilingual \cite{devlin2018bert} &  & 97.78 & 75.95 \\ \hline
MiniLM-L12-v2 \cite{reimers-2019-sentence-bert} & \SetCell[r=2]{c}{Sentence Similarity} &  98.14 & 76.52 \\ 
MiniLM-L12-v2-multilingual \cite{reimers-2019-sentence-bert} &  &  97.76 & \textbf{79.79} \\ \hline  
\end{tblr}
\caption{\small{Experimental results on our pill dataset for the text-image matching task using CLIP model with different BERT pretrained models. Here, CNN model is \texttt{MobileNet-V3-small}. Best results are highlighted in \textbf{bold}.}}
\vspace{-15pt}
\label{tab:clip_experiment}
\end{table}
\textbf{\\ \\CLIP Model.} We use the SOTA model in text-image matching called Contrastive Language–Image Pre-training (CLIP)~\cite{radford2021learning} as the second baseline model in this research. It has been proposed to learn visual concepts with language supervision. We train CLIP model on the pill image dataset with different pre-trained language models (e.g., BERT \cite{devlin2018bert}), while the vision model is the \texttt{MobileNet-V3-small}, which has the best results on baseline classification. The results for Scenario 1-1 are shown in Table~\ref{tab:clip_experiment}. The results indicate that the pre-trained language models in different tasks (e.g., Fill-Mask, Sentence Similarity) are almost similar. The best test set result for the CNN model trained from scratch is $79.79\%$.

Based on the experimental results reported in Tables \ref{tab:experiment_cnn} and \ref{tab:clip_experiment}, we chose to use the trained from scratch CNN model \texttt{MobileNet-V3-small} and the pre-trained language model \texttt{MiniLM-L12-v2-multilingual} for our following experiments and comparisons.

\subsubsection{Comparison with Baseline Approaches}
\textbf{\\ \\Scenario 1.} Table~\ref{tab:exp_scenario_1} summarizes the results of the proposal PIMA and the baseline models concerning Scenarios 1-1, 1-2, and 1-3. For all scenarios, PIMA achieves the highest F1-score of $98.88\%$, $98.41\%$, and $89.77\%$, respectively. Even without using a GNN, PIMA w/o Graph still outperforms the others. Specifically, it outperforms the baseline classification by $23.94\%$, $29.10\%$, and $34.01\%$, respectively.
In comparison with CLIP, PIMA improves the F1-scores by $19.09\%, 25.78\%$ and $41,60\%$ concerning Scenarios 1-1, 1-2, and 1-3, respectively. 
We observe that the proposed model achieves a high level of performance even when the number of pill images and prescription samples in Scenarios 1-2 and 1-3 are very limited.
\begin{table}[tb]
\begin{tblr}{
  colspec = {p{3.5cm}|X[c]|X[c]|X[c]},
}
\hline
\SetCell[r=2]{l}{\textbf{Model}} & \SetCell[c=3]{c}{\textbf{F1-Score (\%)}} \\ \cline{2-4} 
& \textbf{Scenario 1-1} & \textbf{Scenario 1-2} &  \textbf{Scenario 1-3} \\ 
\hline
Baseline & 71.65 & 66.07 & 42.82 \\  
CLIP & 79.79 & 72.63 & 48.17 \\ 
PIMA w/o Graph & 95.59 & 95.17 & 76.83 \\ 
PIMA (our proposal) & \textbf{98.88} & \textbf{98.41} & \textbf{89.77} \\\hline
\end{tblr}
\caption{\small{Experiment results concerning Scenario 1 on our pill image dataset. PIMA significantly outperforms other SOTA methods. Best results are highlighted in \textbf{bold}.}}
\label{tab:exp_scenario_1}
\end{table}

\begin{table}[tb]
\begin{tblr}{
  colspec = {p{3cm}|X[c]X[c]X[c]|X[c]X[c]X[c]},
}
\hline
\SetCell[r=2]{l}{\textbf{Model}} & \SetCell[c=3]{c}{\textbf{Scenario 2-1}} &&& \SetCell[c=3]{c}{\textbf{Scenario 2-2}}\\
\cline{2-7} 
& \SetCell[c=1]{c}{F1(Cor) {\fontsize{8pt}{8pt}\selectfont (\%)}} & \SetCell[c=1]{c}{F1(Mis) {\fontsize{8pt}{8pt}\selectfont (\%)}} &
\SetCell[c=1]{c}{F1(Avg) {\fontsize{8pt}{8pt}\selectfont (\%)}} &
\SetCell[c=1]{c}{F1(Cor) {\fontsize{8pt}{8pt}\selectfont (\%)}} & 
\SetCell[c=1]{c}{F1(Mis) {\fontsize{8pt}{8pt}\selectfont (\%)}} &
\SetCell[c=1]{c}{F1(Avg) {\fontsize{8pt}{8pt}\selectfont (\%)}} \\ \hline
CLIP & 77.47 & 31.42 & 54.46 & 82.81 & 54.84 & 68.83 \\ 
PIMA w/o Graph & 96.11 & 30.06 & 63.09 & 97.24 & 55.45 & 76.35 \\ 
PIMA (our proposal) & \textbf{96.85} & \textbf{62.10} & \textbf{79.48} & \textbf{97.99} & \textbf{94.35} & \textbf{96.17} \\ \hline
\end{tblr}
\caption{\small{Matching accuracy concerning Scenario 2.}}
\label{tab:exp_02}
\end{table}
\textbf{\\Scenario 2.}
In this experiment, we set the threshold to consider an image and text pair matching correctly as $\alpha$ (with $\alpha=0.8$ for all experimental scenarios). 
To ease the presentation, we denote F1(Cor) as the F1-score for matching pills that correctly taken with theirs corresponding names in the prescription.
Besides, we use F1(Mis) to indicate the F1-score for detecting pills that are not in the prescription, and F1(Avg) to represent the mean of F1(Cor) and F1(Mis). 
The results of Scenarios 2-1 and 2-2 are shown in Table~\ref{tab:exp_02}. 
As shown, PIMA achieves the highest average accuracy in both scenarios, $79.48\%$ and $96.17\%$, respectively. 
Specifically, PIMA’s F1(Cor) is higher than that of CLIP by $19.38\%$ and $15.18\%$, respectively. 
In addition, PIMA improves the accuracy of detecting the use of wrong pills by $30.68\%$ in Scenario 2-1 and $39.51\%$ in Scenario 2-2 compared to PIMA. 
In comparison between F1(Cor) in the two scenarios, it can be seen that F1(Cor) in Scenario 2-1 is lower than that in Scenario 2-2 because the mistaken pills in Scenario 2-1 is much more similar to the prescribed pills.
Finally, it can be observed that PIMA w/o Graph achieves a higher F1(Cor) than CLIP. However, the F1(Mis) of PIMA w/o Graph is similar to that of CLIP's. 

\vspace{-5pt}
\subsection{Discussion}
According to the experimental results, PIMA outperforms all other approaches in terms of both accuracy and convergence speed.
As stated in the previous section, PIMA enhanced the F1 score in terms of accuracy from $19.05\%$ to $46.95\%$ compared to comparison benchmarks.
These results are the consequence of contrastive learning's impacts.
In fact, the baseline model separates pill and text box recognition into two distinct phases. The error in pill-prescription matching is then accumulated by the sum of the errors generated by the two phases.
For the CLIP model, the matching loss only considers matched pairs. Specifically, it only intends to minimize the distance between pill images and pill names. This technique creates a paradox in the circumstance of pills with extremely similar appearances but significantly different names.
Specifically, in such a case, the two similar visual representations are pulled back to nearly two completely different textual representations.
In contrast to baseline and CLIP, PIMA employs contrastive loss, which takes into account both matched and mismatched pairs of pill names and images; additionally, we use weights while computing contrastive loss to balance the contribution of matched and mismatched pairs. This weighted contrastive loss increases the model's generalizability and avoids it from being skewed toward the mismatched cases. Thus, PIMA achieves higher precision.

\begin{figure}[tb]
  \begin{subfigure}[b]{0.45\textwidth}
    \includegraphics[width=\textwidth]{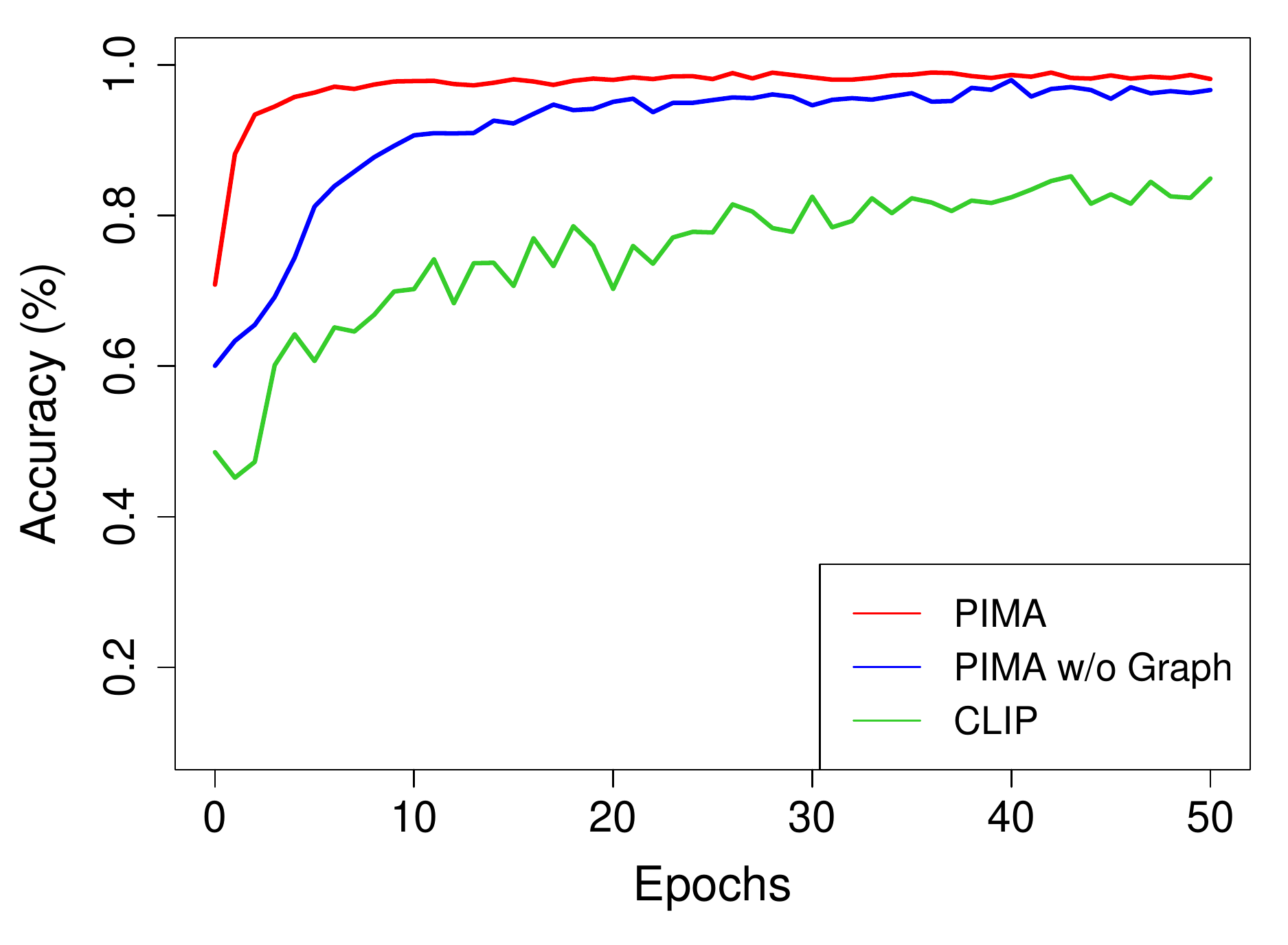}
    \caption{\small{Accuracy for matching correctly taken pills}}
    \label{fig:f1}
  \end{subfigure}
  \hfill
  \begin{subfigure}[b]{0.45\textwidth}
    \includegraphics[width=\textwidth]{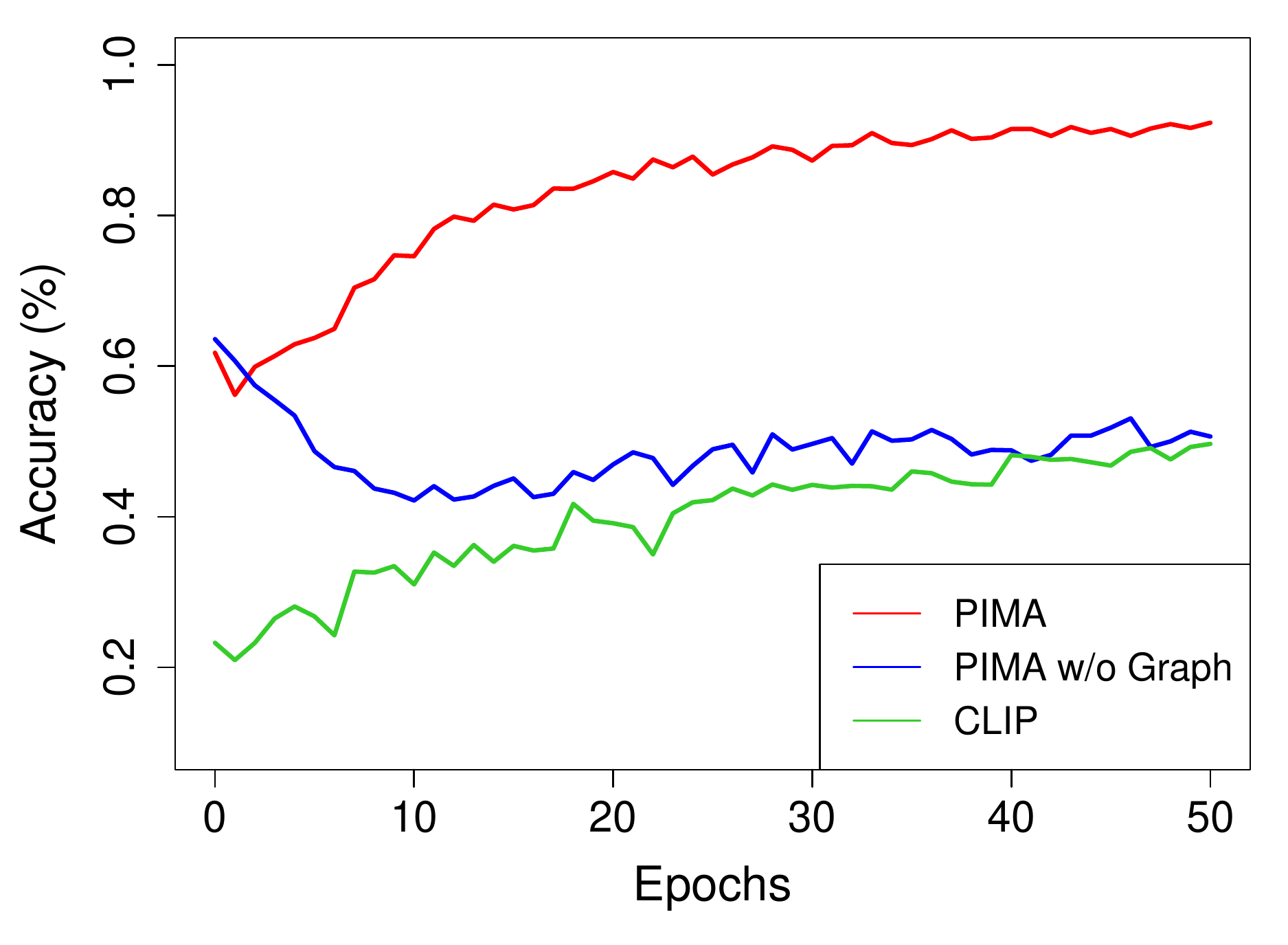}
    \caption{\small{Accuracy for detecting mistaken pills}}
    \label{fig:f2}
  \end{subfigure}
  \caption{\small{Convergence speed on Scenario 2-2.  \label{fig:traing_loss}}\vspace{-10pt}}
\end{figure}

Regarding convergence speed, Fig.~ \ref{fig:f1} demonstrates that after only $50$ epochs, PIMA has converged to an accuracy of about $97\%$ for matching correctly taken pills. In contrast, CLIP only achieves an accuracy of approximately $80\%$.
Although PIMA w/o Graph achieves significantly better performance than CLIP, its accuracy after $10$ epochs is only about $88\%$, whereas that of PIMA is about $95\%$.
Fig.~ \ref{fig:f2} depicts the convergence rate for the detection of mistaken pills.
It is apparent that PIMA outperforms the other two significantly.
In particular, after $50$ epochs, PIMA achieves an accuracy of approximately $90\%$, whereas CLIP and PIMA w/o Graph only reach $50\%$. 
This can be explained by the GNN-based pseudo classifier's contribution.
By having this classifier highlight text boxes containing medicine names, we have narrowed the search space for the matching issue, allowing PIMA to converge significantly more quickly than comparison benchmarks.

\vspace{-5pt}
\section{Conclusion}
\label{sec:conclusion}
We presented PIMA, a novel method to solve the pill-prescription matching task. The key idea behind the PIMA learning framework is the use of a Graph Neural Network (GNN) architecture and contrastive learning to jointly learn text and image representations in order to enhance pill-prescription matching performance. Our extensive experiments on a real-world pill dataset, including pill and prescription images, show that the proposed approach significantly outperforms baseline approaches, enhancing the matching F1-score from $19.05\%$ to $46.95\%$. Additionally, we also demonstrated that the proposed PIMA is able to achieve a high level of performance while requiring less training costs compared to other benchmarks. We release our code at \url{https://github.com/AIoT-Lab-BKAI/PIMA}.

\section*{Acknowledgement}
This work was funded by Vingroup Joint Stock Company (Vingroup JSC), Vingroup, and supported by Vingroup Innovation Foundation (VINIF) under project code VINIF.2021.DA00128.

%
%
%
\bibliographystyle{splncs04}
\bibliography{ref}

\begin{thebibliography}{10}
\providecommand{\url}[1]{\texttt{#1}}
\providecommand{\urlprefix}{URL }
\providecommand{\doi}[1]{https://doi.org/#1}

\bibitem{8679954}
Chang, W.J., Chen, L.B., Hsu, C.H., Lin, C.P., Yang, T.C.: A deep
  learning-based intelligent medicine recognition system for chronic patients.
  IEEE Access  \textbf{7},  44441--44458 (2019)

\bibitem{deng2009imagenet}
Deng, J., Dong, W., Socher, R., Li, L.J., Li, K., Fei-Fei, L.: {ImageNet}: A
  large-scale hierarchical image database. In: Proceedings of the 2009 IEEE
  Conference on Computer Vision and Pattern Recognition. pp. 248--255 (2009)

\bibitem{devlin2018bert}
Devlin, J., Chang, M.W., Lee, K., Toutanova, K.: {BERT}: Pre-training of deep
  bidirectional transformers for language understanding. Computation Research
  Repository arXiv Preprint, arXiv:1810.04805  (2018)

\bibitem{dosovitskiy2020image}
Dosovitskiy, A., Beyer, L., Kolesnikov, A., Weissenborn, D., Zhai, X.,
  Unterthiner, T., Dehghani, M., Minderer, M., Heigold, G., Gelly, S.,
  Uszkoreit, J., Houlsby, N.: An image is worth 16x16 words: Transformers for
  image recognition at scale. Computation Research Repository arXiv Preprint,
  arXiv:2010.11929  (2020)

\bibitem{fromedeep}
Frome, A., Corrado, G.S., Shlens, J., Bengio, S., Dean, J., Ranzato, M.A.,
  Mikolov, T.: {DeViSE}: A deep visual-semantic embedding model. Neural
  Information Processing Systems  \textbf{26},  2121–2129 (2013)

\bibitem{gao2020fashionbert}
Gao, D., Jin, L., Chen, B., Qiu, M., Li, P., Wei, Y., Hu, Y., Wang, H.:
  {FashionBERT}: Text and image matching with adaptive loss for cross-modal
  retrieval. Computation Research Repository arXiv Preprint, arXiv:2005.09801
  (2020)

\bibitem{hamilton2017inductive}
Hamilton, W., Ying, Z., Leskovec, J.: Inductive representation learning on
  large graphs. Neural Information Processing Systems  \textbf{30},
  1025–1035 (2017)

\bibitem{he2016deep}
He, K., Zhang, X., Ren, S., Sun, J.: Deep residual learning for image
  recognition. In: Proceedings of the 2016 IEEE Conference on Computer Vision
  and Pattern Recognition. pp. 770--778 (2016)

\bibitem{hendrycks2016gaussian}
Hendrycks, D., Gimpel, K.: Gaussian error linear units ({GELUs}). Computation
  Research Repository arXiv Preprint, arXiv:1606.08415  (2016)

\bibitem{hochreiter1997long}
Hochreiter, S., Schmidhuber, J.: Long short-term memory. Neural Computation
  \textbf{9},  1735--1780 (1997)

\bibitem{DBLP:journals/corr/abs-1905-02244}
Howard, A., Sandler, M., Chu, G., Chen, L., Chen, B., Tan, M., Wang, W., Zhu,
  Y., Pang, R., Vasudevan, V., Le, Q.V., Adam, H.: Searching for {MobileNetV3}.
  Computation Research Repository arXiv Preprint, arXiv:1905.02244  (2019)

\bibitem{kiros2014unifying}
Kiros, R., Salakhutdinov, R., Zemel, R.: Unifying visual-semantic embeddings
  with multimodal neural language models. Computation Research Repository arXiv
  Preprint, arXiv:1411.2539  (2014)

\bibitem{loshchilov2017decoupled}
Loshchilov, I., Hutter, F.: Decoupled weight decay regularization. Computation
  Research Repository arXiv Preprint, arXiv:1411.2539  (2019)

\bibitem{radford2021learning}
Radford, A., Kim, J.W., Hallacy, C., Ramesh, A., Goh, G., Agarwal, S., Sastry,
  G., Askell, A., Mishkin, P., Clark, J., Krueger, G., Sutskever, I.: Learning
  transferable visual models from natural language supervision  \textbf{139},
  8748--8763 (2021)

\bibitem{reimers-2019-sentence-bert}
Reimers, N., Gurevych, I.: {Sentence-BERT}: Sentence embeddings using {S}iamese
  {BERT}-networks. In: Proceedings of the 2019 Conference on Empirical Methods
  in Natural Language Processing. pp. 3973--3983 (2019)

\bibitem{sandler2018mobilenetv2}
Sandler, M., Howard, A., Zhu, M., Zhmoginov, A., Chen, L.C.: {MobileNetV2}:
  Inverted residuals and linear bottlenecks. In: Proceedings of the 2018 IEEE
  Conference on Computer Vision and Pattern Recognition. pp. 4510--4520 (2018)

\bibitem{sechidis2011stratification}
Sechidis, K., Tsoumakas, G., Vlahavas, I.: On the stratification of multi-label
  data. In: Proceedings of the 2011 European Conference on Machine Learning and
  Knowledge Discovery in Databases. vol.~3, pp. 145--158 (2011)

\bibitem{vaswani2017attention}
Vaswani, A., Shazeer, N., Parmar, N., Uszkoreit, J., Jones, L., Gomez, A.N.,
  Kaiser, L.u., Polosukhin, I.: Attention is all you need. Advances in Neural
  Information Processing Systems  \textbf{30} (2017)

\bibitem{8010584}
Yaniv, Z., Faruque, J., Howe, S., Dunn, K., Sharlip, D., Bond, A., Perillan,
  P., Bodenreider, O., Ackerman, M., Yoo, T.: The national library of medicine
  pill image recognition challenge: An initial report. In: Proceedings of the
  2016 IEEE Applied Imagery Pattern Recognition Workshop (2016)

\end{thebibliography}

\end{document}